\documentclass[10pt,twocolumn,letterpaper]{article}

\usepackage{tikzit}

\usepackage{cvpr}
\usepackage{times}
\usepackage{epsfig}
\usepackage{graphicx}
\usepackage{amsmath}
\usepackage{amssymb}
\usepackage{float}
\usepackage{hhline}
\usepackage[toc,page]{appendix}
\usepackage{url}

\graphicspath{ {./img/} }

\usepackage{hyperref}
\hypersetup{breaklinks=true,bookmarks=false}

\cvprfinalcopy 

\def\cvprPaperID{****} 

\tikzstyle{green dot}=[fill={rgb,255: red,11; green,255; blue,39}, draw=black, shape=circle, line width=3pt]
\tikzstyle{red dot}=[fill={rgb,255: red,255; green,36; blue,7}, draw=black, shape=circle]
\tikzstyle{dense layer box}=[fill=white, draw=black, shape=rectangle, inner sep=0pt, minimum height=0.5cm, minimum width=1cm]
\tikzstyle{circle}=[fill=white, draw=black, shape=circle]
\tikzstyle{small circle}=[fill=white, draw=black, shape=circle, inner sep=0pt, minimum size=0.25cm, tikzit draw=cyan]
\tikzstyle{small circle (concat operation)}=[fill=white, draw=black, shape=circle, inner sep=0pt, minimum size=0.4cm, tikzit draw=yellow]

\tikzstyle{left arrow}=[->]
\tikzstyle{dashed}=[-, dotted]
\tikzstyle{invisible}=[-, draw=white]

\begin{document}

\title{
    \hrule height 4pt
    \vskip7mm
    From Spectral Graph Convolutions to Large Scale \\Graph Convolutional Networks \\
    \hrulefill}

\author{Matteo Bunino\\
EURECOM\\
{\tt\small matteo.bunino@eurecom.fr}

}

\maketitle

\begin{abstract}
  Graph Convolutional Networks (GCNs) have been shown to be a powerful concept that has been successfully applied to a large variety of tasks across many domains over the past years. 
  In this work we study the theory that paved the way to the definition of GCN, including related parts of classical graph theory.
  We also discuss and experimentally demonstrate key properties and limitations of GCNs such as those caused by the statistical dependency of samples, introduced by the edges of the graph, which
  causes the estimates of the full gradient to be biased. Another limitation we discuss is the negative impact of minibatch sampling on the model performance. As a consequence, during parameter update, gradients are computed on the whole dataset, undermining scalability to large graphs. To account for this, we research alternative methods which allow to safely learn good parameters while sampling only a subset of data per iteration.
  We reproduce the results reported in the work of Kipf et al. \cite{Kipf2016} and propose an implementation inspired to SIGN \cite{frasca2020}, which is a sampling-free minibatch method.
  Eventually we compare the two implementations on a benchmark dataset, proving that they are comparable in terms of prediction accuracy for the task of semi-supervised node classification. 
\end{abstract}

\section{Introduction}
Graph Convolutional Networks (GCNs) have become the \textit{de facto} standard baseline architecture for graph neural networks and inspired many follow-up works in the field of machine learning on graphs, including semi-supervised vertices labeling \cite{Kipf2016}, link prediction \cite{zhang2018link}, anomaly detection \cite{jiang_anomaly_2019, mesgaran_graph_2020} and other domains \cite{xia2021}.
Unlike other Artificial Neural Networks, GCNs are based on deeper theoretical bases that come from the spectral graph theory, originally developed to address the task of graph clustering. In fact, graphs can be seen as generalizations of regular manifolds, thus some operations require an \textit{ad-hoc} mathematical generalization to the graph domain.
The operation performed by GCNs is the convolution of a graph signal (or function) with a learnable filter. This means bringing into a vertex additional information from its neighbors according to some rule, leveraging the structural information encoded by the graph. 
For Convolutional Neural Networks (CNNs), the convolution of a signal (image) with a filter results in another tensor where each element is a linear combination of its neighbors and itself.
While it is straightforward to define convolution among vectors or matrices (e.g. in CNNs), in the graph setting this requires further mathematical derivations.

Graph convolutions can be implemented in the spatial domain, as GraphSAGE \cite{hamilton2017}. However, this approach is computationally expensive and very difficult to scale to large graphs.
Alternatively, graph convolutions can be performed in the Fourier (spectral) domain. Spectral convolutions require less computations, after some mathematical adaptations. GCN \cite{Kipf2016} is the result of the approximation of spectral graph convolutions as a deep neural network, whose parameters are trained with gradient descent to minimize a well-defined cost function.

Designing solutions capable of scaling to huge graphs is an active field of research, as there is a strong interest of applying deep learning  to those settings where historically it has been prohibitively expensive. Some examples include social networks and financial graphs, which are characterized by a significantly large number of vertices. Applying graph convolutional networks to financial graphs could allow to spot financial frauds or complex money laundering schemes, hardly noticeable otherwise. Thus, also governments and other institutions are interested in applying machine learning to huge graphs.

Traditionally, deep learning models scale to datasets of million of examples by iteratively updating the trainable parameters considering only a small subsets of data at each step. Minibatches of data are randomly sampled from a uniform distribution over the whole dataset and the gradients computed before each backpropagation are an unbiased estimate of the full gradient (i.e. obtained when considering the whole dataset). This property holds as long as the samples in the dataset are independent and identically distributed, \textit{i.i.d.} for short.
In contrast, graph data intrinsically share some statistical dependencies represented by edges, which invalidates the i.i.d. assumption. Furthermore, uniformly sampling nodes from a graph produces a new graph, degrading the original structural information. As a consequence, the GCN model \cite{Kipf2016} circumvents this problem by employing the full gradient, being hardly scalable to graphs containing millions or billions of nodes due to memory constraints.

The contribution of this work is threefold:
\begin{itemize}
    \item First, we perform the study of the theoretical background of GCN and a portion of the literature concerning graph convolutional networks, including the works of \cite{bruna2014, deferrard2016, hammond2011}. 
    This includes the in-depth study of the theory behind spectral graph convolutions, concerning the graph Laplacian, spectral graph clustering and graph Fourier transform, which are respectively discussed in Appendices \ref{appendix_graph_laplacian}, \ref{appendix_spectral_clustering} and \ref{appendix_graph_fourier}.
    \item Second, we analyze how graph neural networks can be applied to huge graphs, explaining why GCN is hardly scalable and proposing a solution inspired by a recent work of \cite{frasca2020}.
    \item Eventually, we replicate the results achieved by GCN \cite{Kipf2016}, comparing it with the SIGN model \cite{frasca2020} on the Cora benchmark dataset.
\end{itemize} 

In Section \ref{method_section} we present the theory that paves the way from spectral graph convolutions to GCN. Then, we perform a literature research regarding the current state of the art of minibatch approaches and graph sampling including \cite{chen2018}, \cite{chiang2019}, \cite{zeng2020} and \cite{frasca2020}. Section \ref{sec_exp} describes the experimental setting, whereas Section \ref{sec_discuss} outlines a discussion concerning the comparison of GCN and SIGN methods and current limitations of graph neural networks.

This work has been developed during a semester project at EURECOM university, under the supervision of Prof. Pietro Michiardi\footnote{\texttt{pietro.michiardi@eurecom.fr}}.

\section{Towards graph neural networks}\label{method_section}
This section outlines the path of successive refinements that derived graph neural networks models like GCN \cite{Kipf2016} and SIGN \cite{frasca2020} from spectral graph convolutions, highlighting the key steps that paved the way to modern graph machine learning.  


\subsection{Spectral graph convolutions}
The following sections will assume an undirected graph $G=(V,\mathcal{E})$ with $N = |V|$ vertices, characterized by an adjacency matrix $A$, a diagonal degree matrix $D_{i,i}=\sum_jA_{i,j}$ and a combinatorial Laplacian $L=D-A$. Alternatively, without loss of generality, it can be used the symmetric normalized Laplacian $L=I_N-D^{-1/2}AD^{-1/2}$, where $I_N$ is the identity matrix.

Spectral graph convolutions were initially introduced by Bruna et al. \cite{bruna2014} where the convolution of a graph signal $\mathbf{x} \in R^N$ with a vector filter $\mathbf{g} \in R^N$ can be formally defined as:
\begin{equation}
    \mathbf{g} \star \mathbf{x} = \mathcal{F}^{-1}\left(\mathcal{F}(\mathbf{g}) \odot \mathcal{F}(\mathbf{x}) \right)
\end{equation}
according to the convolution theorem. Note that $\odot$ represents the Hadamard product.
According to graph Fourier transform theory presented in Appendix \ref{appendix_graph_fourier}, we have
\begin{equation}
\label{spectral_conv}
   \begin{aligned}
    \mathbf{g} \star \mathbf{x} &= U\left(U^T\mathbf{g} \odot U^T\mathbf{x} \right)\\
     &= U\left(\widehat{\mathbf{g}} \odot U^T\mathbf{x} \right)\\
    & =  U\left(\widehat{G} \cdot U^T\mathbf{x} \right)
   \end{aligned}
\end{equation}
where $U$ is the matrix of the Laplacian eigenvectors and $\widehat{G} \in R^{N\times N}$ is a diagonal matrix with the vector $\widehat{\mathbf{g}} = U^T\mathbf{g} \in R^N$ on its diagonal, introduced to rewrite the Hadamard product between two vectors as a dot product between a diagonal matrix and a vector.

According to Equation (\ref{spectral_conv}) and referring to \cite{bruna2014}, it is now possible define the propagation rule where each layer $\psi=1\dots \Psi$  transforms an input
vector $\mathbf{x}_\psi$ of size $N \times f_{\psi-1}$ into an output $\mathbf{x}_{\psi+1}$ of dimensions  $N \times f_{\psi}$ as:
\begin{equation}
\label{multi_layer_spect_convo}
\mathbf{x}_{\psi+1,j} = \sigma\left(U\sum_{i=1}^{f_{\psi-1}}\widehat{G}_{\psi,i,j} U^T \mathbf{x}_{\psi,i}\right), 
\; j=1\dots f_\psi
\end{equation}
Where $\sigma(\cdot)$ is a real non-linearity and $\widehat{G}_{\psi,i,j}$ is a diagonal $N\times N$ matrix of parameters to learn.
It is easily noticeable from the previous equation that for a single layer $\psi$, it is required to learn $f_{\psi-1} \cdot f_\psi \cdot N$ parameters.

\subsection{Localization of filters}
\label{filters_localiz}
The drawback of Equations (\ref{spectral_conv}) and (\ref{multi_layer_spect_convo}) is that they encode the same behavior as spectral convolution for regular grids, hence the learnable filters $\widehat{g}$ are global and their size depends on the number of vertices. Furthermore, large filters are often difficult to train since they are not guaranteed to identify useful patterns. 
As explained in Appendix \ref{appendix_graph_laplacian}, as the magnitude of the eigenvalues increase, the Dirichlet energy of the corresponding eigenvectors increases as well, meaning fewer smoothness. This can be seen as the analogous of higher frequency Fourier modes that compose a signal.

\begin{figure}[t]
\begin{center}
   \includegraphics[width=1\linewidth]{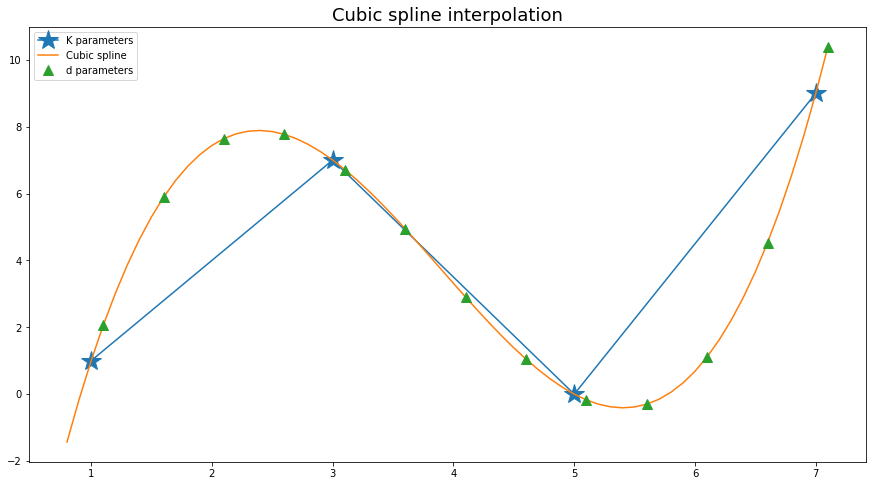}
\end{center}
   \caption{The blue star markers represent the original $K$ parameters, the orange line is the cubic spline interpolation and the green triangle markers represent the $d$ points resulting from the interpolation. It is visually straightforward that the new vector of $d$ parameters is smoother that the original one of $K$ parameters.}
\label{fig:long}
\label{fig:cubic_spline}
\end{figure}

To reduce the number of parameters, \cite{bruna2014} analyze two possible solutions.
The first consist of performing the analogous of a low-pass filtering on the graph Fourier basis, by taking only the first $d$ eigenvectors, which is justified by the fact that lower frequencies components encode most of the useful information whereas the higher frequencies modes are roughly spurious noise. $U$ can be rewritten as $U_d \in R^{N\times d}$ and the number of parameters for each layer is reduced to $f_{\psi-1} \cdot f_\psi \cdot d$.
The shortcoming of this approach is that $d$ requires tuning and it is still true that $d = O(N)$. Furthermore, \cite{bruna2014} explains how it is important to learn filters with a localized scope in the spatial domain.
To address both of these problems, the authors propose to learn a fixed number $K$ of parameters and reconstruct a vector of $d$ parameteres $\widehat{\mathbf{g}} \in R^d$ resorting to a spline interpolation as follows:
\begin{equation}
\label{cubic_spline}
    diag(\widehat{G}_{\psi,i,j}) \simeq \mathcal{K}\cdot\alpha_{\psi,i,j}
\end{equation}
where $\mathcal{K}$ is a $d\times K$ fixed cubic spline kernel
\cite{noauthor_interpolating_nodate} and $\alpha_{\psi,i,j}$ are the $K$ spline coefficients. A visual representation of cubic spline interpolation is shown in Figure \ref{fig:cubic_spline}.
This interpolation has two benefits: 
\begin{itemize}
    \item It enables to reconstruct $d$ elements from $K\ll d$ elements, reducing the parameters to be learned for each filter to $f_{\psi-1} \cdot f_\psi \cdot K$, which is constant in the graph size.
    Note that $\mathcal{K}$ is fixed and does not have to be learned.
    \item Cubic splines introduce a smoothing on the parameteres vector $ diag(\widehat{G}_{\psi,i,j})$, namely in the spectral domain.
\end{itemize}
According to spectral theory, performing a smoothing in the spectral domain results into an increased localization in the spatial domain. In other words, the filters that we are learning are still global, but their effectiveness in the spatial domain fades out with distance.  

\subsection{Chebyshev polynomials}
While Equation (\ref{cubic_spline}) makes the learned filter independent from the graph size, it still requires the eigendecomposition of the Laplacian matrix which has a complexity of $O(N^3)$ and costly multiplication with the eigenvectors matrix $U$ with complexity $O(N^2)$, which prevent this method from being scalable to large graphs. 
Furthermore, when evaluating on different graphs, we would like to exclude the structure of the training graph encoded by eigenvectors.
These are two good reasons that led Hammond et al. \cite{hammond2011} and Deferrard et al. \cite{deferrard2016} to further improve Equation (\ref{cubic_spline}) with the introduction of Chebyshev polynomials.
As explained in \cite{deferrard2016}, Equation (\ref{spectral_conv}) can be rewritten as 
\begin{subequations}
    \begin{alignat}{2}
    \mathbf{y} &= \mathbf{\theta} \star \mathbf{x} \\
    &= g_\theta(L)\,\mathbf{x} \label{g_l_filter}\\
    &= g_\theta(U\Lambda U^T)\,\mathbf{x} = Ug_\theta(\Lambda)U^T\mathbf{x} \label{g_lambda_filter}
    \end{alignat}
\end{subequations}
where $\theta \in R^N$ is a filter, $g_\theta(\Lambda) = diag(\widehat{\theta})$, being $\widehat{\theta}$ the Graph Fourier Transform (GFT) of $\theta$. In Equation (\ref{g_l_filter}), the convolution is expressed as a filtering operation treating $g_\theta(L)$ as a transfer function, a well known formulation in signal processing theory.
However, $g_\theta(\Lambda)$ can also be polynomially parametrized by a linear combination of $K$ powers of $\Lambda$
\begin{equation}
    g_\theta(\Lambda) = \sum_{k=0}^{K-1}\theta_k\Lambda^k
\end{equation}
with a similar idea of Bruna et al. \cite{bruna2014}, but now $g_\theta$ being a function of the eigenvalues of $L$.
The main goal of \cite{deferrard2016} is to express the graph convolution as in Equation (\ref{g_l_filter}) rather than like in Equation (\ref{g_lambda_filter}), in order to avoid the complexity of matrix multiplications.
It turns out that this can be done by parametrizing $g_\theta(L)$ as a polynomial function that can be computed recursively from $L$. Note that $K$ multiplications of a sparse $L$ have complexity $O(K|\mathcal{E}|) \ll O(N^2)$. 
One possible polynomial is the Chebyshev expansion, that in \cite{hammond2011} is used to approximate wavelets kernel.
In fact, Chebyshev polynomials can be expressed recursively, as $T_k(\mathbf{x}) = 2\,\mathbf{x}\,T_{k-1}(\mathbf{x}) - T_{k-2}(\mathbf{x})$ with $T_0(\mathbf{x}) = 1$ and $T_1(\mathbf{x}) = \mathbf{x}$.
\begin{equation}
\label{cheby_filter}
    g_\theta(L) = \sum_{k=0}^{K-1}\theta_kT_k(\Tilde{L})
\end{equation}
where $\Tilde{L}=2L/\lambda_{max}-I_N$ is the rescaled version of $L$, in order to rescale its eigenvalues in $[-1,1]$ and $\theta_k$ is the $k$-th component of parameters vector $\theta$. $T_k(\Tilde{L})$ is fixed and does not have to be learned: to speedup the training phase, the Chebyshev polynomials can be precomputed since they only depend on the Laplacian.
The filtering operation of Equation (\ref{g_l_filter}) can be expressed as:
\begin{equation}
\label{cheby_propagation}
    \mathbf{y} = \sum_{k=0}^{K-1}\theta_kT_k(\Tilde{L})\cdot \mathbf{x}
\end{equation}
Thus, the propagation rule of Equation (\ref{multi_layer_spect_convo}) where each layer $\psi=1\dots \Psi$  transforms an input
vector $\mathbf{x}_\psi$ of size $N \times f_{\psi-1}$ into an output $\mathbf{x}_{\psi+1}$ of dimensions  $N \times f_{\psi}$ becomes:
\begin{equation}
\label{multi_layer_cheby}
\mathbf{x}_{\psi+1,j} = \sigma\left(\sum_{i=1}^{f_{\psi-1}}\mathbf{g}_{\theta;\psi,i,j}(L) \cdot \mathbf{x}_{\psi,i}\right), 
\; j=1\dots f_\psi
\end{equation}
As a result, the model proposed in the work of Deferrard et al. \cite{deferrard2016} does not further reduce the number of parameters for each layer with respect to Bruna et al. \cite{bruna2014}, which is always $f_{\psi-1} \cdot f_\psi \cdot K$, but it achieves better performances and generalization to other graphs by avoiding to compute the expensive products with the eigenvectors matrix $U$. Recall that eigenvectors encode the structure of the graph on which the model is trained.
As explained in \cite{deferrard2016}, the polynomial parametrization of $g_\theta(L)$ with Chebyshev polynomials of order $k$ hides $k$-th powers of the graph Laplacian, which has an important semantic meaning. In fact, it can be proven that $L^k=U\Lambda^kU^T$ is the Laplacian of $k$-hop neighbors with respect to a central vertex, hence $T_k(\Tilde{L})$ now only depends on vertices that are $k$ steps far from a given vertex. As a consequence, $g_\theta(L)$ expressed as in Equation (\ref{cheby_filter}) only depends on vertices that are at a maximum $K-1$ steps from a central vertex. This is another way to achieve filters localization, previously motivated in Section \ref{filters_localiz}.

\subsection{Graph convolutional networks (GCN)}
\label{gcn}
The work of Kipf et al. \cite{Kipf2016} proposes a solution to tackle two of the main drawbacks of the method introduced by \cite{deferrard2016}:
\begin{itemize}
    \item The necessity to compute the eigenvalues of the Laplacian matrix, achievable through eigendecomposition with complexity $O(N^3)$. Or alternatively compute the powers of the Laplacian.
    \item Each layer $\psi=1\dots\Psi$ has a number of trainable parameters equal to $f_{\psi-1} \cdot f_\psi \cdot K$. Reducing the number of parameters may have the benefit of increasing the ability of generalization and speedup the convergence of the optimizer.   
\end{itemize}
The authors propose to limit the layer-wise convolution in Equation (\ref{cheby_propagation}) to $K=2$, namely performing a convolution with up to 1-hop neighbors and approximating $\lambda_{\max} \approx 2$:
\begin{equation}
\label{kipf_conv_1}
\begin{aligned}
     \mathbf{x} \star \mathbf{g} &\approx \theta_0\,\mathbf{x}+\theta_1(L-I_N)\,\mathbf{x} \\
     &\approx\theta_0\,\mathbf{x}-\theta_1D^{-1/2}AD^{-1/2}\,\mathbf{x}
\end{aligned}
\end{equation}
where $L=I_N-D^{-1/2}AD^{-1/2}$ and $\theta_i$ is a parameter.
The parameters are reduced by setting $\theta_0=-\theta_1=\theta$ and Equation (\ref{kipf_conv_1}) is simplified as
\begin{equation}
\label{param_reduct}
    \mathbf{x} \star \mathbf{g} \approx \theta \, (I_N + D^{-1/2}AD^{-1/2}) \, x
\end{equation}
It can be proven that $I_N + D^{-1/2}AD^{-1/2}$ has eigenvalues bounded in $[0,2]$.
To prevent numerical instabilities (vanishing or exploding gradients) during the model training, a \textit{renormalization trick} is implemented as follows: 
$I_N + D^{-1/2}AD^{-1/2} \rightarrow \Tilde{D}^{-1/2}\Tilde{A}\Tilde{D}^{-1/2}$
. Where $\Tilde{A} = A+I_N$ and $\Tilde{D}_{i,i} = \sum_{j}\Tilde{A}_{i,j}$.
The propagation rule in Equation (\ref{multi_layer_cheby}) can be further rewritten as
\begin{equation}
\label{multi_layer_kipf}
X_{\psi+1} = \sigma\left(\Tilde{D}^{-1/2}\Tilde{A}\Tilde{D}^{-1/2} \, X_\psi \, \Theta_\psi \right)
\end{equation}
where $X_{\psi+1} \in R^{N\times f_\psi}$, $X_{\psi} \in R^{N\times f_{\psi-1}}$ and $\Theta_\psi \in R^{f_{\psi-1}\times f_{\psi}}$.
This convolution has now complexity of $O(|\mathcal{E}|\,f_{\psi-1}\,f_{\psi})$ since the product $(\Tilde{D}^{-1/2}\Tilde{A}\Tilde{D}^{-1/2}) \cdot X_\psi$ can be efficiently implemented as a product of a sparse matrix with a dense matrix.
As a result, each layer performs a convolution taking into account only the 1-hop neighbors of a central vertex employing half as much as the parameters of \cite{deferrard2016}, as explained in Equation (\ref{param_reduct}). Furthermore, since at each layer only Chebyshev polynomials of order 0 and 1 are employed, there is no more need to compute the eigendecomposition of $L$ and powers of $L$.
To perform convolutions up to the $n$-hop neighbor, the authors propose to stack $\Psi$ layers in order to approximate the effect of a single Chebyshev layer with $K=\Psi+1$.
This is justified in Deep Learning literature where it is known that the composition of locally linear functions (hidden layers with ReLU activations) can approximate well any non linear function.
The authors of \cite{Kipf2016} present a study on the model depth, $\Psi$, by adding residual connections (He at al. \cite{kaiming2016}) at each layer
\begin{equation}
    X_{\psi+1} = \sigma\left(\Tilde{D}^{-1/2}\Tilde{A}\Tilde{D}^{-1/2} \, X_\psi \, \Theta_\psi \right)
    +  X_\psi
\end{equation}
For the datasets considered by the authors, the best results have been obtained with 2- or 3-layered models. Hence, the convolution up to 2- or 3-hop neighbors conveys most of the meaning, underlining the importance of learning spatially localized filters. Therefore, the depth of a GCN is a critical parameter which has to be tuned and it influences the complexity of patterns we are trying to learn from the graph. Furthermore, the depth of a GCN is bounded to the structure of a graph -- to have a meaning, it should be upper-bounded by the longest path on the graph.


\subsection{Minibatch sampling methods}
As discussed by \cite{Kipf2016}, when using full gradient approach, memory requirement grows linearly in the size of the dataset.
It is hence necessary to implement a minibatch strategy in order to make this method scalable to large graphs.
In the semi-supervised vertex classification setting, the vertices play the role of samples on which the model is trained. Most of traditional machine learning is based on the assumption of samples being \textit{independently} drawn from the same distribution, which allows to decompose the loss into the independent contribution of each sample. This paves the way to well-known optimization techniques as Stochastic Gradient Descent (SGD).
However, this is not true in the graph setting where the samples (vertices) are inter-related by edges, that result in creating statistical dependence. Thus, sampling a minibatch of data-points both introduces sampling \textit{bias} and it degenerates the original graph structure.

GraphSAGE \cite{hamilton2017} follows the intuition discussed in Section \ref{gcn} under which, in order to compute the training loss on a GCN having $\Psi$ layers, only the $\Psi$-hop neighbours are necessary.
However, this is not always enough and there can be graphs (e.g. social networks) in which the number of $k$-hop neighbors grows exponentially in $k$, not fitting in memory.
For this reason, GraphSAGE recursively samples uniformly with replacement up to $N$ neighbors for each $k$-hop neighbor. This procedure ensures to upper bound the number of sampled neighbors for a vertex to $O(N^\Psi)$. Then, if the batch contains $B$ vertices, the batch size in memory will be $O(B\,N^\Psi)$.
This method has two main drawbacks which entail a considerable waste of computational resources:
\begin{itemize}
    \item Sampling with replacement performed on a graph with loops can introduce redundant neighbors.
    \item At each iteration, the loss is computed on $B$ vertices of the batch, although the convolutions were performed on $O(B\,N^\Psi)$ vertices.
\end{itemize}

To tackle the problem of redundant computation, other methods like ClusterGCN \cite{chiang2019} and GraphSAINT \cite{zeng2020} were introduced, which perform \textit{graph-sampling} instead of GraphSAGE's \textit{neighbourhood-sampling}. In graph-sampling approaches, a batch is obtained as a sampled sub-graph of the original graph. However, the sub-graph has to preserve a meaningful structure, similar to the original graph.

ClusterGCN \cite{chiang2019} performs a clustering on the graph and employs the clusters as batches. At each iteration the parameters of the GCN model are updated according to a cluster's information. This has the benefit of perfectly preserving the original graph structure, being as much connected as possible. However, this is subject to the overhead of graph clustering, as discussed in Appendix \ref{appendix_spectral_clustering}.

GraphSAINT \cite{zeng2020}, on the other hand, proposed a sampling model that can be implemented according to different schemes: uniform vertex sampling, uniform edge sampling, or \textit{importance sampling} by using random walks to compute the importance of vertices and use it as the probability distribution for sampling.

FastGCN \cite{chen2018} addresses the problem of bias introduced by sampling statistically correlated vertices. The authors propose an alternative formulation of graph convolutions, defined as integral transforms of embedding functions under probability measures. This formulation assumes that the vertices are i.i.d. and that each layer of the network defines an embedding function of vertices. Both the loss and the gradients are obtained by evaluating the integral transforms of embedding functions via Monte Carlo simulation.
With this expedient, the obtained batch loss is a \textit{consistent} estimator of the true loss and the model can be optimized with standard SGD.

\subsubsection{Scalable graph neural networks (SIGN)}
\label{sign_pres}
Frasca et al. \cite{frasca2020} propose SIGN, a sampling-free method which is a promising alternative to the methods described in the previous section.
They argue that it is still unclear whether sampling provides positive effects other than just reducing resources requirements, thus proposing a sampling-free alternative to address huge graphs. Furthermore, the implementation of sampling schemes introduces non-negligible overheads and according to the Occam's razor principle, a simpler approach is often preferable.
\begin{figure*}[t]
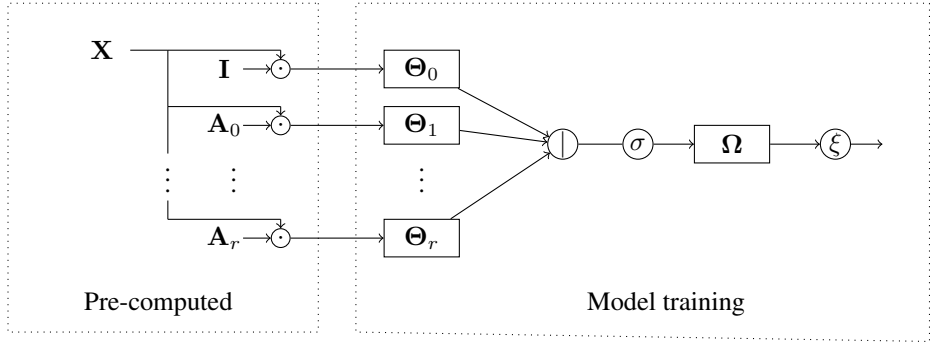


\begin{center}
\ctikzfig{img/sign}
\end{center}
   \caption{SIGN architecture. $\Theta_k$ is the $k$-th dense layer which transforms vertex features after they have been aggregated according to $A_k$ and $\Omega$ represents the dense layer used to compute final predictions after the features resulting from the previous layer have been concatenated ($|$). The $\cdot$ operation indicates the matrix products $\mathbf{A}_i\cdot\mathbf{X}$.}
\label{fig:long}
\label{fig:onecol}
\end{figure*}
The SIGN architecture is composed of two layers. Graph convolutions are computed in the first layer, where multiple fixed neighbour aggregators are applied to each vertex and the resulting features are processed in parallel by a set of dense layers, one for each aggregator. The results of each dense layer are concatenated and sent to the second layer, represented by another dense layer, which computes the final predictions.
Fixed neighbors aggregations can be pre-computed since they do not depend on the network's parameters.
As suggested by the authors, pre-aggregation can be efficiently carried out resorting to  well-known frameworks for big data processing, such as Apache Spark, alleviating the cost of training.
Once neighbors aggregation is pre-computed, minibatches of data can be safely sampled without violating samples independence assumption or degrading the structure of the graph, thus allowing for scalable minibatch training.
The effect of the $k$-th neighborhood aggregator is to define a rule to convolve the features of the $k$-hop neighbors of a central vertex.
In practice, the aggregator $A_k$ can be considered as an ordinary adjacency matrix employed in a GCN convolution operation.

The authors argue that SIGN is capable of achieving performances comparable to GraphSAINT \cite{zeng2020} and ClusterGCN \cite{chiang2019} but better than FastGCN \cite{chen2018}. Moreover, this method has proved to be way faster at training and inference time with respect to GraphSAINT and ClusterGCN. Encouraged by these achievements we decided to implement SIGN, comparing its performances with GCN.
The authors of SIGN \cite{frasca2020} propose a combination of different approaches to aggregate neighbors information, including simple, PPR-based and triangle-based adjacency matrices. In this work we opt for the simplest approach, being more easily comparable with GCN. We employ matrix powers $k=0\dots K-1$ of the graph adjacency matrix as aggregators, defined as follows:
\begin{equation}
    A_k = \Tilde{D}(k)^{-1/2} \: \Tilde{A}(k) \: \Tilde{D}(k)^{-1/2}
\end{equation}
where $\Tilde{A}(k) = A^k + I_N$ and $ \Tilde{D}_{i,i}(k) = \sum_j \Tilde{A}_{i,j}(k)$ and $A_0=I_N$.
An interesting property of an adjacency matrix $A$ is that $A^n$ exposes $n$-hop connections between vertices. Without loss of generality, $A(k)$ is an adjacency matrix.
The addition of the identity matrix to $A^k$ is equivalent to add a self loop to each vertex, which in practice allows to aggregate its features with its $k$-hop neighbors. 
The normalization after the matrix power is required in order to avoid numerical instability during training, as explained by \cite{Kipf2016}.
The model can be expressed as:
\begin{equation}
    \begin{aligned}
    Z &= \sigma\left(X\Theta_0 \, | \, A_1X\Theta_1 \, | \, \dots \, | \, A_rX\Theta_r\right)\\
    Y &= \xi(Z\Omega)
    \end{aligned}
\end{equation}
where $\sigma(\cdot)$ is the activation function (e.g. ReLU), $\xi(\cdot)$ is the Softmax activation, $X$ is the $N\times F_0$ features matrix, $\Theta_i$ is a $F_0\times F_1$ parameters matrix for convolution with $k$-hop neighbors and $\Omega$ is a $(r+1)\,F_1\times C$ hidden layer matrix of the final fully connected layer.

\section{Experiments}\label{sec_exp}
The goals of this work encompass the reproduction of the empirical results presented by Kipf et al. \cite{Kipf2016} and the exploration of improvements over the vanilla GCN, better suited for large graphs. To this end, this section presents a comparison between the results obtained with GCN \cite{Kipf2016} and SIGN \cite{frasca2020} models, for the task of semi-supervised node classification on Cora dataset.
For the sake of interpretability, all the experiments have been carried out on the same train, validation, and test dataset partitions. 

\subsection{Reproduction of GCN results on Cora}
\begin{figure*}
\begin{center}
   \includegraphics[width=1\linewidth, height=7cm]{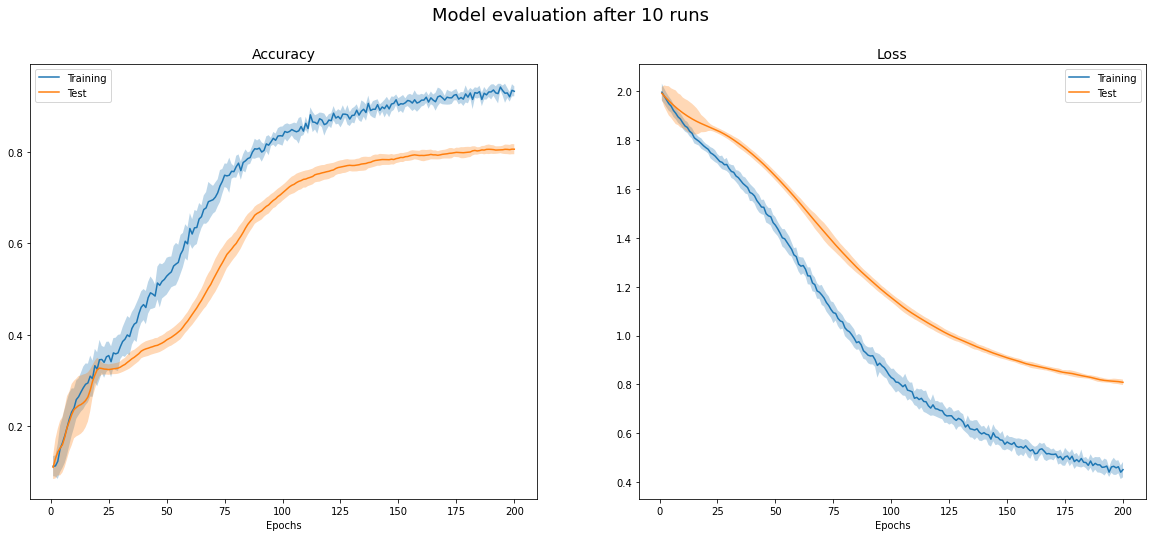}
\end{center}
   \caption{GCN learning curves with error bars for 10 different random initializations, on Cora dataset. Train (blue), test (orange).}
\label{fig:long}
\label{gcn_10_fig}
\end{figure*}

\begin{figure*}
\begin{center}
   \includegraphics[width=1\linewidth, height=7cm]{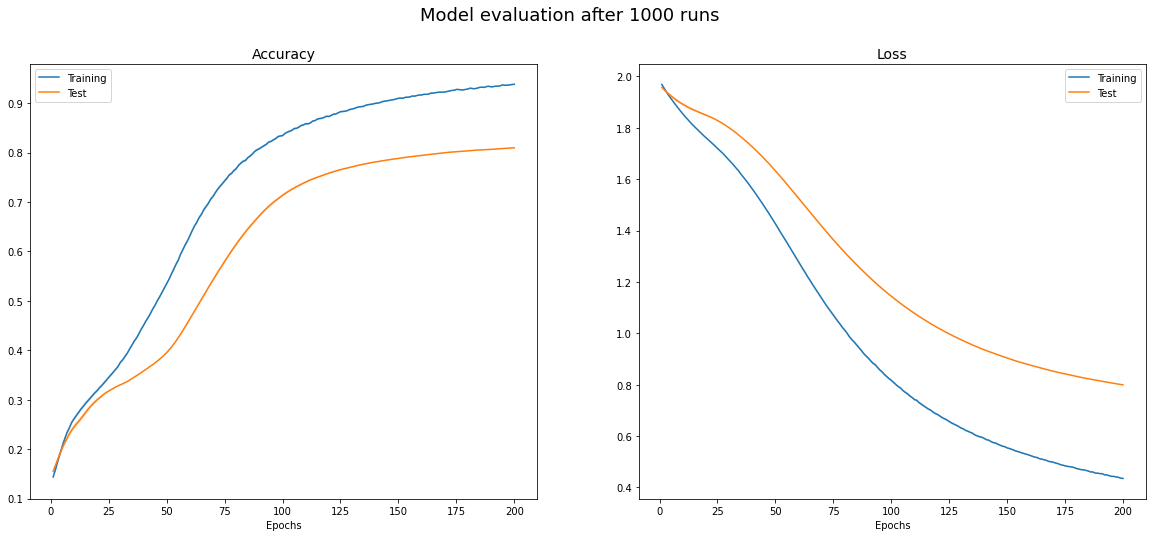}
\end{center}
   \caption{GCN learning curves with error bars for 1K different random initializations, on Cora dataset. Train (blue), test (orange).}
\label{fig:long}
\label{gcn_1k_fig}
\end{figure*}

The results originally achieved by the GCN model \cite{Kipf2016} were obtained with a TensorFlow implementation. In this work, we develop an alternative implementation based on the PyTorch framework. Achieving similar results underlines the model robustness and its implementation invariance. The GCN model is evaluated on the Cora dataset as in \cite{Kipf2016}.
The training is carried out with the hyperparameters reported in Table \ref{gcn_hyperparams}.
\begin{table}
\begin{center}
\begin{tabular}{|l|c|}
\hline
Hyperparameter & Values \\
\hline\hline
Optimizer & Adam\\
Learning rate & 0.01\\
Weight decay & 5e-4\\
Layers & 2\\
Hidden units & 16\\
Dropout & 0.5\\
Layers init. & xavier \cite{xavier}\\
Epochs & 200\\
\hline
\end{tabular}
\end{center}
\caption{Best hyperparameters used for GCN training.}
\label{gcn_hyperparams}
\end{table}
To give a statistical meaning to the results, in our experiments we re-trained the GCN model using 1K different random seeds, thus reporting 95\% confidence intervals for the evaluation metrics.
The results of our implementation are compared with the ones of \cite{Kipf2016} in Table \ref{gcn_results}. Figure \ref{gcn_10_fig} and Figure \ref{gcn_1k_fig} show the learning curves with 95\% confidence intervals.
\begin{table}
\begin{center}
\begin{tabular}{|l|c|c|c|}
\hline
Model & Test accuracy & Train time (s) & Reruns\\
\hline\hline
Our GCN & $81.0 \pm 0.1$ & $1.165 \pm 0.003$ & 1K \\
GCN \cite{Kipf2016} & 81.5 & 4 & -\\
\hline
\end{tabular}
\end{center}
\caption{Results for GCN model and comparison with reference paper \cite{Kipf2016}. Results on Cora dataset.}
\label{gcn_results}
\end{table}


\subsection{Minibatch on SIGN model}
\begin{figure*}
\begin{center}
   \includegraphics[width=1\linewidth, height=7cm]{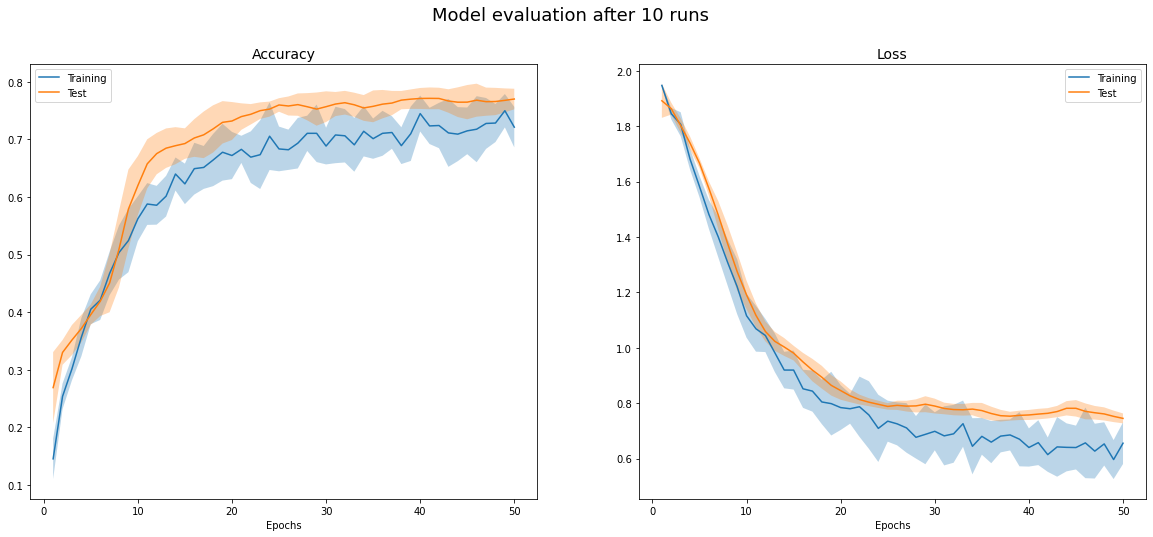}
\end{center}
   \caption{SIGN learning curves with error bars after 10 train repetitions on Cora. Train (blue), test (orange).}
\label{fig:long}
\label{sign_10_fig}
\end{figure*}

\begin{figure*}
\begin{center}
   \includegraphics[width=1\linewidth, height=7cm]{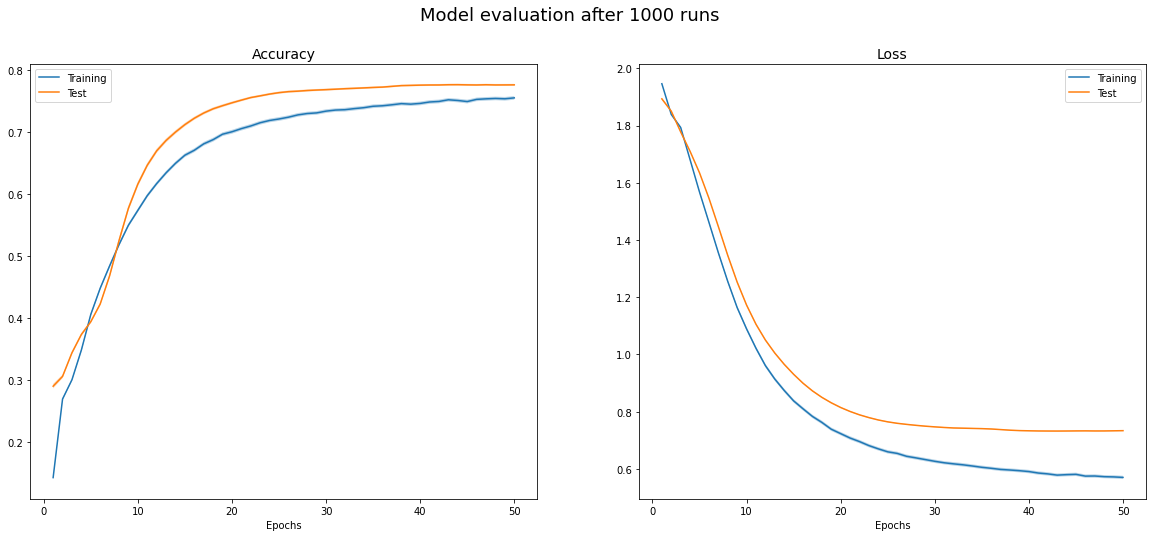}
\end{center}
   \caption{SIGN learning curves with error bars after 1k train repetitions on Cora. Train (blue), test (orange).}
\label{fig:long}
\label{sign_1k_fig}
\end{figure*}

The hyperparameters which showed the most critical importance during tuning are: learning rate, number of aggregators and the number of hidden units. Surprisingly, the batch size has not influenced a lot the outcome of training. Adam optimizer performed better in that SGD in all the experiments: a possible justification is the critical role played by the learning rate. In fact, Adam implements an adaptive strategy, whereas the SGD uses a fixed step size when updating the parameters.
The number of parameters in the last layer $\Omega$ directly depends on the number $h$ of hidden units of each dense layer applied during graph convolutions, as $O(h\,(r+1))$. Therefore, controlling the number of hidden units is key to prevent overfitting.
In addition, to regularize the model we introduced dropout for each convolutional layer $\Theta_i$ and in the final fully connected layer $\Omega$.
The best hyperparameters found during an informal search are reported in Table \ref{sign_hyperparams}.
The results of this model are reported in Table \ref{sign_results} with 95\% confidence intervals, while Figure \ref{sign_10_fig} and Figure \ref{sign_1k_fig} show the learning curves with error bars.
\begin{table}
\begin{center}
\begin{tabular}{|l|c|}
\hline
Hyperparameter & Values \\
\hline\hline
Optimizer & Adam\\
Learning rate & 0.15\\
Weight decay & 1e-5\\
Batch size & 512 \\
Aggregators ($r$) & 4\\
Hidden units & 8\\
Dropout & 0.5\\
Layers init. & xavier \cite{xavier}\\
Epochs & 50\\
\hline
\end{tabular}
\end{center}
\caption{Best hyperparameters used for SIGN training.}
\label{sign_hyperparams}
\end{table}
\begin{table}
\begin{center}
\begin{tabular}{|l|c|c|c|}
\hline
Model & Test accuracy & Train time (s) & Reruns\\
\hline\hline
SIGN & $77.7 \pm 0.1$ & $2.165 \pm 0.005$ & 1K \\
\hline
\end{tabular}
\end{center}
\caption{Results for SIGN model on Cora.}
\label{sign_results}
\end{table}


\section{Discussion}\label{sec_discuss}
The results obtained on the Cora dataset with the GCN model reproduce well the results declared by Kipf et al. \cite{Kipf2016}.
On the other hand, the authors of SIGN \cite{frasca2020} report an improvement over GCN \cite{Kipf2016}, which we were not able to replicate in our experiments. 
Similarly to GCN, we defined neighbor aggregators as adjacency matrices $A_n$, obtained as powers of the original adjacency matrix, whereas the authors of SIGN resorted to sophisticated combinations of different neighbor aggregator approaches. This may justify the discrepancy in the results we obtained. 
Our design choice allows to compare the quality of graph convolutions from a level playing field, employing similar aggregation schemes.

GCN is likely better at learning informative patterns from local neighbors, as the signal is propagated through intermediate neighbors before reaching a node. Instead, in our implementation SIGN works with a coarser approximation of the graph and the information of $n$-th hop neighbors is directly aggregated in a vertex, without the typical intermediation of GCN models. More advanced neighbor aggregation schemes introduced by \cite{frasca2020} may alleviate this problem allowing for better performances, at the cost of requiring domain experts to design neighbors aggregators.

The training time of the SIGN model is on average slower with respect to the GCN model, which in our experiments requires roughly 54\% of the average SIGN training time. 
This may be justified by the fact that SIGN model has a larger number of trainable parameters with respect to GCN. Furthermore, minibatching may introduce a slight overhead due to sampling and memory management, when a new batch is loaded into GPU memory. In fact, GCN performs a single gradient descent update per training epoch, whereas SIGN samples multiple batches of data, performing a backpropagation for each one.  
The parameters of GCN are updated less frequently, which translates to an higher number of epochs to attain convergence. In fact, GCN converges after about 200 training epochs whereas SIGN only requires 50, as emerges when comparing the horizontal axes of Figure \ref{gcn_1k_fig} and Figure \ref{sign_1k_fig}.

The number of neighbor aggregators of the SIGN model is related to the depth of GCN, as both approximate the width of a graph convolutional filter.
While tuning SIGN hyperparameters, the number of aggregators $r$ proved to be a critical for classification accuracy. To reach good performances, SIGN required deeper convolutions in the graph with respect to GCN, taking into account farther neighbors. Likely, wider convolutions compensate for the rougher approximation introduced by SIGN which directly connects a node with its $n$-hop neighbors, rather than mediating the signal by intermediate nodes.

\subsection{Limitations and future work}
Although achieving outstanding results in a large number of tasks, the theory of graph convolutional networks is based on the assumption of symmetric adjacency matrices.
In principle, this restricts the application of GCN-like models to undirected graphs, which is a strong requirement hard to meet in practice. To circumvent this problem, authors proposed some methods to \textit{symmetrize} adjacency matrices. Nevertheless, these are not definitive solutions and the semantics of some graphs may be considerably affected by this approximation. Future work should revisit the bedrock theory on which GCN are based, proposing alternative methods capable of better leveraging directional information in graphs.
Taking as an example the domain of financial graphs, spectral graph theory is not immediately applicable since the transactions define a directed graph where the direction of an edge (e.g. sender and recipient of money) brings some important meaning.
From a mathematical perspective, it is difficult to generalize GCN to directed graphs because, as extensively presented in Appendix \ref{appendix_graph_fourier}, the adjacency matrix of a directed graph is no more guaranteed to be symmetric, therefore a graph Fourier transform is not guaranteed to exist. This undermines the theoretical basis on top of which GCN are built.
A promising solution to train GCN on directed graphs is presented by \cite{ma2019}, which is a good reference for future work. The authors of this work discuss how to symmetrize the adjacency matrix reducing the loss of information, by means of a Perron vectors obtained as left eigenvector of the row-normalized adjacency matrix. According to the proposed implementation, this method may also be applicable to the SIGN minibatch model, by symmetrizing the sampled adjacency matrix identified by a batch, plus some neighbors of each vertex in the batch.

Furthermore, many real-world graphs are actually \textit{dynamic}. For instance, financial graphs evolve over time as new transactions are executed, thus a graph neural network should also learn temporal patterns. This topic constitutes itself a broad research domain, which is left for future work.

\section{Conclusions}
Graph Convolutional Networks presented by Kipf et al. \cite{Kipf2016} introduced a valuable approximation of spectral graph convolutions, allowing to define a new class of neural networks which operate on graph manifolds. Their contribution is the last of a series of works which bridge the gap between spectral graph convolutions and modern graph neural networks. In this work we presented the evolution of graph neural networks and its links with classic spectral graph clustering, underlying the mathematical implications of such evolution. This work is intended to help researchers to understand the roots of graph convolutional networks, when trying to overcome their limitations.

Although GCN has signed a significant breakthrough in the machine learning literature, it comes with its own shortcomings. In particular, GCN is hardly scalable to large graph which cannot fit in memory all at once, as it updates the network parameters using the full gradient computed on the whole dataset. To account for this, we performed a literature research examining graph-sampling and sampling-free methods, identifying a promising solution to scale GCN-like models to huge graphs. 

Eventually, we implement both GCN \cite{Kipf2016} and SIGN \cite{frasca2020} models and compare their performances for the task of semi-supervised node classification on the Cora benchmark dataset, identifying a trade-off between prediction accuracy and scalability of GCN-like models.


 {\small
\bibliographystyle{siam}
\bibliography{bibliography}
 }

\begin{appendices}

%
%

\section{Graph Laplacian matrix}
\label{appendix_graph_laplacian}
This section describes the intuition behind the graph Laplacian matrix, motivating its key role in spectral graph clustering and spectral graph convolutions.

\subsection{The Laplace operator}
Let a function $f: R^n \xrightarrow{} R$, the continuous Laplacian operator of $f$ is defined as:
\begin{align}
    \Delta f &= \sum_i \frac {\partial ^{2}}{\partial x_{i}^{2}}f \\
    &= tr(Hessian(f))=tr(\nabla^2f)
\end{align}
In physics, the continuous Laplacian operator $\Delta$ can be encountered in many applications such as in potential theory or fields theory and an interesting example is the heat equation:
\begin{equation}
    \begin{aligned}
    \frac {\partial u}{\partial t}&={\frac {\partial ^{2}u}{\partial x_{1}^{2}}}+\cdots +{\frac {\partial ^{2}u}{\partial x_{n}^{2}}}\\
    &=\Delta u
\end{aligned}
\end{equation}
which describes how the heat \textit{propagates} through a given region in time.
In fact, following this analogy, we can consider a graph such as a set of points at different temperatures which are connected by a medium represented by the edges where the information flow is represented by heat.
However, generalizing the Laplacian to graphs is also justified by many interesting mathematical properties that will be discussed in details in the next sections.

\subsection{The discrete Laplacian operator}
The Laplacian operator can be defined on a discrete domain by resorting to the finite differences approximation of the derivatives. The centered finite differences for the second derivative can be defined by taking a sufficiently small $h$ and the canonical basis vectors $\{e_1, e_2,\dots e_n\}$:
\begin{equation}
    \frac {\partial ^{2}}{\partial x_{i}^{2}}f(x) \simeq \frac{f(x+e_ih) - 2f(x) + f(x-e_ih)}{h^2}
\end{equation}
Hence, the discrete Laplace operator:
\begin{equation}
\label{discrete_laplace}
    \Delta f \simeq \sum_{i=1}^n \frac{f(x+e_ih) - 2f(x) + f(x-e_ih)}{h^2}
\end{equation}
which is defined on a regular lattice of side equal to $h$.
In $R^2$ this is known as a five-point stencil.
Still on a regular lattice of side $h$, the equation (\ref{discrete_laplace}) can be further simplified as
\begin{equation}
\label{discr_laplace_gen}
    \Delta f(i) \simeq \sum_{i\sim j} \frac{f(j) - f(i)}{h^2}
\end{equation}
where we sum on all neighbors of the point $i$.

\subsection{Graph Laplacian}
A graph $G=(V,\mathcal{E})$ with $N = |V|$ vertices, is a generalization of a regular lattice, where each vertex can have up to $N-1$ neighbors. Furthermore, the function $\phi: V \xrightarrow{} R^{N}$ defined on the graph associates each vertex to a scalar. According to equation (\ref{discr_laplace_gen}) idea and to graph theory, the graph Laplacian of $\phi$ is defined
\begin{equation}
    \label{graph_lapl_1}
    \begin{aligned}
    L\cdot\phi(i) &= \sum_{i\sim j} a_{i,j} [ \phi(i) - \phi(j) ]\\
    &\simeq -\Delta\cdot\phi(i)
    \end{aligned}
\end{equation}
where now $1/h^2$ is substituted by another multiplicative constant $a_{i,j}$ which is an element of the adjacency matrix $A\in R^{N\times N}$. Regarding this term, the graph Laplacian share the same semantic meaning of the discrete Laplace operator when the adjacency matrix encodes the meaning of \textit{closeness} among neighbors.
However, the graph Laplacian is broadly known in its matrix form which can be derived from the Equation (\ref{graph_lapl_1}).
\begin{equation}
    \begin{aligned}
    L\cdot\phi(i) &= \sum_{j} a_{i,j} [ \phi(i) - \phi(j) ]\\
    &= \phi(i) \sum_{j} a_{i,j} - \sum_{j} a_{i,j}\phi(j)\\
    &= \phi(i)\,deg(i) -  \sum_{j} a_{i,j}\phi(j)\\
    &= \sum_{j}(\delta_{i,j}\,deg(i) - a_{i,j})\,\phi(j)\\
    &= \sum_{j}l_{i,j}\,\phi(j)\\\\
    L\phi &= (D-A)\phi
    \end{aligned}
\end{equation}
Where $\phi(i)$ is the $i$-th element of the function $\phi$ and $D_{i,i} = \sum_j a_{i,j}$ is the degree matrix.
We obtain the un-normalized graph Laplacian $L = D-A$.
Another way to derive the Laplace operator is to define it as the divergence of the gradient of a function $f$.
In the graph domain, exploiting again the finite differences approach, the gradient of a function defined on an edge $e=(u,v)$ is
\begin{equation}
    \begin{aligned}
    (\nabla\cdot\phi)_e &= \sqrt{a_{u,v}} [\phi(u)-\phi(v)] \\
    \nabla\cdot \phi &= K^T\phi
    \end{aligned}
\end{equation}
where $K\in R^{N\times|\mathcal{E}|}$ is defined in literature as the incidence matrix. This is also called \textit{edge derivative} in \cite{Shumany2013}.
Thus, the divergence of the gradient for a given vertex $v$
\begin{equation}
    \begin{aligned}
    ( div(\nabla\cdot \phi) )_v &= \sum_{u\sim v} \sqrt{a_{u,v}} [\phi(u)-\phi(v)]\\
    div(\nabla\cdot \phi) &= KK^T\phi\\
    L\phi &= KK^T\phi
    \end{aligned}
\end{equation}
Hence $L = KK^T$.
As explained in \cite{Shumany2013}, an insightful mathematical interpretation of the Laplace operator is given by the Dirichlet energy that describes the \textit{smoothness} of a function in terms of how variable a function is. Let $f: \Omega \subseteq R^n \xrightarrow{} R$,:
\begin{equation}
    {\displaystyle E[f]={\frac {1}{2}}\int _{\Omega }\|\nabla f(x)\|^{2}\,dx,}
\end{equation}

The Dirichlet energy of a function defined on a graph becomes
\begin{equation}
    \begin{aligned}
    E[\phi] &= \frac{1}{2}\sum_{u\sim v} a_{u,v} [\phi(u)-\phi(v)]^2\\
    &= ||K^T\phi||^2 = \phi^TL\phi
    \end{aligned}
\end{equation}
The smoothest function that can be applied on a graph is the solution of the optimization problem
\begin{equation}
    \label{graph_dirichlet}
    \min_\phi E[\phi] = \min_\phi \phi^TL\phi
\end{equation}
Which can be is the Rayleigh quotient for unit norm $\phi$.
Solving the Rayleigh quotient problem
\begin{equation}
    \label{ray_lapl}
    \begin{aligned}
    &\min_{\phi_i} \frac{\phi_i^TL\phi_i}{\phi_i^T\phi_i}\\
    &\text{subject to:}\\
    & \phi_i^T\phi_j = \delta_{i,j}\\
    \end{aligned}
\end{equation}
we obtain a set of orthonormal eigenvectors of the graph Laplacian matrix $\{ \phi_0, \dots, \phi_{n-1} \}$. The corresponding eigenvalues $\lambda_i$ are obtained evaluating the Rayleigh quotient at $\phi_i$. As a result of Equation (\ref{ray_lapl}), the eigenvector $\phi_0$ associated to the smallest eigenvalue is a constant function for all vertices, having all components equal to 1.

%
%

\section{Spectral graph clustering}
\label{appendix_spectral_clustering}
Further developing the theory presented in the previous section it is possible to address the topic of spectral graph clustering and partitioning.
Since the Laplacian matrix is symmetric and positive semidefinite, its eigenvectors form an orthonormal basis and its eigenvalues are all real and nonnegative. Furthermore, for each eigenvalue, its algebraic multiplicity is equal to the geometric multiplicity.
It can be proved from Equation (\ref{ray_lapl}) that the smallest eigenvalue is $\lambda_0=0$, with multiplicity equal to the connected components of the graph \cite{hammond2011}. Hence, if the graph is connected, we can sort the eigenvalues of the Laplacian as:
\begin{equation}
    0 = \lambda_0 < \lambda_1 \leq \lambda_1 \leq \dots \leq \lambda_{n-1}
\end{equation}
The eigenvectors associated to $\lambda_0$ define the connected components and can be represented by vectors having identical components, equal to a constant $\alpha$, that identify vertices belonging to a connected component and 0 everywhere else. If the graph is connected, then the $\phi_0$ is unique and can be expressed as $\phi_0=(1/\sqrt{n}) \, \mathbf{1}_n$, as discussed also in \cite{bruna2014}.
However, in spectral graph clustering we are interested in identifying an arbitrary $K$ number of partitions on a graph, namely finding those $K$ communities of vertices that are well connected among themselves while being poorly connected with the other vertices. In other words, a good cluster has a low \textit{conductance} with the others, as introduced by \cite{Shi1997}.
Given two set of vertices A and B, their conductance is defined as:
\begin{equation}
    C(A,B) = \frac{cut(A,B)}{\min(vol(A), col(B))}
\end{equation}
where:
\begin{itemize}
    \item cut(A,B) is the number of edges between A and B.
    \item vol(A) is the total weighted degree of the vertices in A and is equal to the number of edge endpoints in A.
\end{itemize}
Let the graph $G=(V,\mathcal{E})$ be a connected graph with $N = |V|$ vertices.
Its smallest eigenvalue is $\lambda_0=0$ with multiplicity 1 and its associated eigenvector is a vector $\phi_0=(1/\sqrt{n}) \, 1_n$ but for simplicity can be represented as an un-normalized vector of ones $\phi_0=\mathbf{1}_n$.
Its second smallest eigenvalue can be obtained by solving:
\begin{equation}
    \begin{aligned}
    \lambda_1 &= \min_{\phi_1} \frac{\phi_1^TL\phi_1}{\phi_1^T\phi_1}\\
    &= \min_{(i,j)\in E} (\phi_1(i) - \phi_1(j))^2=\phi_1^TL\phi_1\\
    &\text{subject to:}\\
    & \phi_1^T\mathbf{1} = \sum_{i=1}^N \phi_1(i)= 0, \,\,\, \text{orthogonal to }\phi_0=\mathbf{1}_N\\
    & || \phi_1 ||^2 = \sum_{i=1}^N \phi_1(i)^2 = 1, \,\,\, \text{unit norm}
    \end{aligned}
\end{equation}
This is the first non-zero eigenvalue and is called \textit{algebraic connectivity} of a graph. It was originally introduced in \cite{fiedler73}.
As a consequence of the orthogonality condition, we are obtaining a good partitioning among the vertices because we are asking that the components of $\phi_1$ will be roughly half positive and half negative, entailing a balanced result. However, to avoid the trivial solution of $\phi_1=\mathbf{0}$ and have a normal vector, we also require $\phi_1$ to be unit norm.
This resulting eigenvector has a remarkable importance in the literature of spectral graph clustering and is often referred to as \textit{Fiedler vector}. 
This vector finds the optimal graph partitioning, minimizing the number of edges among different partitions also avoiding to have degenerate partitions such as \{(1), (N-1)\} vertices.
However, the eigenvectors of the graph Laplacian are defined on $R^{N}$, hence their components do not encode a labeling for the clusters. A naive approach would be use the $sign(\phi_i)$ as a labeling function, but there exist more advanced strategies out of the scope of this work, like \cite{Luxburg2007}.

The intuition behind $\lambda_i$ is that the fewer the edges between the two partitions, the smaller the the eigenvalue $\lambda_i$.
Furthermore, as showed in Equation (\ref{graph_dirichlet}), the magnitude of an eigenvalue is also proportional to the smoothness of the relative eigenvector. 
Note that as the multiplication of $\phi$ by a constant does not change the Rayleigh quotient, we can assume without loss of generality that $\phi$ is a unit vector, therefore Equation (\ref{ray_lapl}) and Equation (\ref{graph_dirichlet}) are equivalent.

The first eigenvector of the graph Laplacian is the smoothest function we can find on the graph. The second eigenvector is the next smoothest function of all graph functions that are orthogonal to the first one and so on.
The first eigenvector of the Laplacian can be seen as the DC component of the graph signal and the next eigenvectors are the higher frequency modes of the signal.
This intuition shows a link between spectral graph clustering and graph Fourier transform.

%
%

\section{Graph Fourier transform}
\label{appendix_graph_fourier}
As stated in the previous section, as long as the connected graph $G=(V,\mathcal{E})$ with $N=|V|$ vertices is undirected, its Adjacency and Laplacian matrices are symmetric. As a consequence, the eigenvectors of the Laplacian can be chosen to form an orthonormal basis.
The spectral theorem states that a symmetric matrix $L$ can be diagonalized as:
\begin{equation}
    L = U\Lambda U^T
\end{equation}
where $U$ is the matrix of the eigenvectors of $L$ and $\Lambda$ has its eigenvalues on the diagonal.
The graph Fourier transform can be defined in analogy with the Fourier transform on the real line, as explained in \cite{hammond2011}.
In $R$, the complex exponentials $e^{j\omega x}$ are eigenfunctions with respect to the Laplacian operator $\frac {d^2}{d x^2}$.
\begin{equation}
    \Delta e^{j\omega x} = \frac {d^2}{d x^2}e^{j\omega x} = -\omega^2e^{j\omega x}
\end{equation}
The Fourier transform of $f:R\xrightarrow{}R$ is defined as
\begin{equation}
    \begin{aligned}
    \widehat{f}(\omega) &= \mathcal{F}[f](\omega) = \frac{1}{2\pi} \langle f(x), e^{j\omega x}\rangle\\
    &= \frac{1}{2\pi}\int_{-\infty}^{\infty}f(x) (e^{j\omega x})^* dx
    \end{aligned}
\end{equation}
which can be seen as a projection of $f(x)$ in a new space defined by the Fourier basis.
On the other hand, $f(x)$ can be written as an expansion in terms of complex exponential in the inverse Fourier transform
\begin{equation}
    \begin{aligned}
    f(x) &= \mathcal{F}^{-1}[\widehat{f}](x) = \frac{1}{2\pi} \langle \widehat{f}(\omega), e^{-j\omega x}\rangle\\
    &= \frac{1}{2\pi}\int_{-\infty}^{\infty}\widehat{f}(\omega) (e^{-j\omega x})^* d\omega
    \end{aligned}
\end{equation}
Analogously, in the graph domain, we can define a graph Fourier basis as the set of orthonormal eigenvectors of the Laplacian matrix. Hence, given a function $\phi(i): V_i \xrightarrow{} R$ defined on the vertices of $G$, namely $\phi \in R^N$, we can define the graph Fourier transform evaluated at an eigenvalue $\lambda_l$ as:
\begin{equation}
    \begin{aligned}
    \widehat{\phi}(\lambda_l) &= \mathcal{GF}[\phi](\lambda_l) = \langle \phi, u_l\rangle\\
    &=\sum_{n=1}^N\phi(n)\,u_l^*(n)
    \end{aligned}
\end{equation}
where $u_l$ is the eigenvector associated to $\lambda_l$.
In matrix form becomes:
\begin{equation}
    \widehat{\phi}(\Lambda) = U^T\cdot\phi
\end{equation}
The inverse transform is:
\begin{equation}
    \begin{aligned}
    \phi(i) = \mathcal{GF}^{-1}[\widehat{\phi}](i) 
    =\sum_{l=0}^{N-1}\widehat{\phi}(\lambda_l)\,u_l(i)
    \end{aligned}
\end{equation}
In matrix form:
\begin{equation}
    \phi = U\cdot \widehat{\phi}(\Lambda)
\end{equation}

\subsection{Similarities with PCA}
If the eigenvectors of the Laplacian can be seen as the elements of the Fourier basis in the graph domain, the magnitude of their associated eigenvalues is a measure of their smoothness (according to Dirichlet energy) and can be interpreted as the analogous of frequency in the Euclidean domain. 
Hence the eigenvectors associated to larger eigenvalues are can be considered as higher frequencies modes.
As discussed in the work of Bruna et al. \cite{bruna2014}, it is often true that most of the useful structural information of a graph is encoded by the $d$ smoothest (lower frequencies) eigenvectors, whereas the other ones contain mostly spurious noise. For this reason, \cite{bruna2014} motivates as beneficial a low-pass filtering by taking only the first $d$ eigenvectors associated with the \textit{lowest} magnitude eigenvalues.
An identical operation, from a mathematical standpoint, is carried out when performing the Principal Component Analysis (PCA) on a dataset $X \in R^{r\times c}$ by performing the eigendecomposition of the sample covariance matrix $\frac{1}{r}(X-\mu_X)^T(X-\mu_X)$, which is symmetric by construction, and taking the $d$ eigenvectors associated with the \textit{highest} magnitude eigenvectors.
In the case of PCA, we can argue that this operation is equivalent to performing an high-pass filtering on the eigenvectors of the sample covariance matrix. This choice is justified by the meaning that assume the eigenvalues in this case: they are proportional to the variance of data along their associated eigenvector direction.
However, in both Graph Fourier Transform and PCA, data of the original domain are projected in a new domain spanned by the eigenvectors of a reference matrix, after some filtering has been applied to them. 

\end{appendices}

\end{document}